\documentclass{article}

\usepackage{arxiv}

\usepackage[utf8]{inputenc} 
\usepackage[T1]{fontenc}    
\usepackage{url}            
\usepackage{booktabs}       
\usepackage{amsfonts}       
\usepackage{nicefrac}       
\usepackage{microtype}      
\usepackage{lipsum}		
\usepackage{graphicx}
\usepackage{natbib}
\usepackage{doi}

\usepackage{times}
\usepackage{soul}
\usepackage{url}
\usepackage[utf8]{inputenc}
\usepackage[small]{caption}
\usepackage{graphicx}
\usepackage{amsmath}
\usepackage{amsthm}
\usepackage{booktabs}
\usepackage{algorithm}
\usepackage{algorithmic}
\usepackage{amsmath}
\usepackage{amssymb}
\usepackage[switch]{lineno}

\newcommand{\N}{N} 
\newcommand{\featureSpace}{\mathbb{F}}
\newcommand{\tagSpace}{\mathbb{B}}
\newcommand{\K}{K} 
 
\newcommand{\D}{D} 
\newcommand{\descrSpace}{\mathbb{D}}
\newcommand{\Part}{\mathbf{P}} 
\newcommand{\Dts}{\mathcal{O}} 
\newcommand{\elem}{o} 
\newcommand{\defff}{\stackrel{def}{=}}

\newcommand{\nbBaseClust}{H} 
\newcommand{\nbCand}{V} 
\newcommand{\nbPat}{A} 
\newcommand{\nbAppar}{\mathbf{S}} 
\newcommand{\patternByClust}{\mathbb{W}} 
\newcommand{\candidateDescr}{\mathbf{Y}}
\newcommand{\instanceByClust}{\mathbb{I}}

\usepackage{multirow}
\usepackage{caption}
\usepackage{subcaption}

\usepackage{color}

\title{Towards Explainable Clustering: A Constrained Declarative based Approach}

\author{Mathieu Guilbert, Christel Vrain, Thi-Bich-Hanh Dao\\
	LIFO\\
	Univ. Orléans, INSA Centre Val de Loire\\\
	Orléans, France \\
	\texttt{name.surname@univ-orleans.fr} \\
}

\renewcommand{\shorttitle}{Towards Explainable Clustering: A Constrained Declarative based Approach}

\hypersetup{
pdftitle={Towards Explainable Clustering: A Constrained Declarative based Approach},
pdfsubject={q-bio.NC, q-bio.QM},
pdfauthor={Mathieu Guilbert},
pdfkeywords={First keyword, Second keyword, More},
}

\begin{document}

\maketitle

\begin{abstract}
The domain of explainable AI is of interest in all Machine Learning fields, and it is all the more important in clustering, an unsupervised task whose result must be validated by a domain expert. We aim at finding a clustering that has high quality in terms of classic clustering criteria and that is explainable, and we argue that these two dimensions must be considered when building the clustering. 
We consider that a good global explanation of a clustering should give the characteristics of each cluster taking into account their abilities to describe its objects (coverage) while distinguishing it from the other clusters (discrimination). Furthermore, we aim at leveraging expert knowledge, at different levels, on the structure of the expected clustering or on its explanations. 
In our framework an explanation of a cluster is a set of patterns, and we propose a novel interpretable constrained clustering method called ECS for declarative clustering with Explainabilty-driven Cluster Selection that integrates structural or domain expert knowledge expressed by means of constraints. 
It is based on the notion of coverage and discrimination that are formalized at different levels (cluster / clustering), each allowing for exceptions through parameterized thresholds. Our method relies on four steps: generation of a set of partitions, computation of frequent patterns for each cluster, pruning clusters that violates some constraints, and selection of clusters and associated patterns to build an interpretable clustering. This last step is combinatorial and we have developed a Constraint-Programming (CP) model to solve it. The method can integrate prior knowledge in the form of user constraints, both before or in the CP model.
\end{abstract}

\section{Introduction} \label{sec:intro}

Clustering is a machine learning process that aims at grouping objects into clusters according to their similarities. 
Clustering is often seen as an exploratory task exploiting unlabelled data, and  it is therefore crucial to understand the clusters in order to validate them. In this paper, we propose a new method for explainable clustering. 
We consider that a good global explanation of a clustering should give the characteristics of each cluster taking into account their abilities to describe the objects of the cluster (coverage) while distinguishing them from the other clusters (discrimination). 
Most of existing explainable clustering methods aim at explaining an existent clustering, using decision trees  \cite{dasgupta2020explainable} or surrounding shapes \cite{chen2016interpretable,lawless2021interpretable}. Some methods construct a clustering where each cluster is described by a pattern \cite{pami-MichalskiS83,mlj-Fisher87,dao2018descriptive}. However no method constructs a good clustering considering both the coverage and discrimination requirements, as well as prior knowledge. Moreover, we consider that to be interesting a cluster must not only group similar objects but also be explainable, and that these two properties should be enforced when building a clustering. In this work, we propose a method that builds a clustering and an explanation for each cluster at the same time, differently from most explainable clustering methods.

Furthermore, prior expert knowledge exists and we aim at leveraging them, at different levels, on the structure of the expected clustering or on its explanations. Thus our work extends the framework of constrained clustering \cite{wagstaff2001constrained,basu2008constrained}. Our framework can integrate classic constraints on cluster composition, such as  \textit{Must-Link (ML)} (resp. \textit{Cannot-Link (CL)}) constraints expressing that two instances must be put into the same (resp. different) cluster(s) \cite{wagstaff2001constrained}, 
constraints on the sizes of clusters or on the margin between clusters \cite{basu2008constrained}.
Moreover, constraints on explanations can also be integrated.

We introduce a new interpretable constrained clustering method called ECS. Data is described by two views, possibly not disjoint: one for clustering and any kind of data can be used, the other is Boolean to build the explanations. For instance, in real-world applications, molecules can be described by reaction values against some tested compounds and binary structural properties.
It first builds a pool of candidate clusters along with a set of covering patterns for each cluster, and subsequently performs cluster and pattern selection to satisfy all the  constraints.
This (combinatorial) step is accomplished using Constraint Programming (CP), a declarative framework allowing the expression of a wide range of constraints.

The contributions of this paper are:

\begin{itemize}
    \item A formalization of the notion of interpretable clustering, through configurable constraints on coverage and discrimination.
    \item A novel CP clustering model to create  a clustering and the explanation of each cluster from a set of clusters and their descriptions.
    \item A clustering based on the generation of a set of base partitions (as in ensemble methods), allowing the selection of clusters of different shapes.
    \item The possibility of taking advantage of expert knowledge at different stages of the cluster selection process.
    \item The introduction of three novel clustering explanation quality measures.
\end{itemize}

The paper is organized as follows: related work (Sec. \ref{sec:sotA}), our formulation of interpretable clustering (Sec. \ref{sec:pbSta}), the method (Sec. \ref{sec:novel}) and the CP model (Sect. \ref{sec:model}), experimental results (Sec. \ref{sec:exp}), and a conclusion (Sec. \ref{sec:conclu}).

\section{Related work} \label{sec:sotA}

The domain of explainable AI (XAI) has attracted a lot of attention in recent years. In clustering many work rely on decision trees (see for instance \cite{izza2022tackling} for a discussion on their interpretability).
Some decision trees are built a posteriori to approximate results of K-Means or K-Medians algorithms \cite{dasgupta2020explainable,frost2020exkmc,gamlath2021nearly,esfandiari2022almost,charikar2022near,laber2021price,makarychev2021explainable,makarychev2021near,laber2023shallow,bandyapadhyay2023find}. 
Explainable-by-design approaches have also been introduced with novel split criteria such as cluster compactness \cite{loyola2020explainable} or the Silhouette and Dunn index \cite{bertsimas2021interpretable}.

Another family of approaches focuses on finding surrounding shapes of the clusters: each cluster in the partition is described by a set of rules delimiting its bounds. Those shapes are hyper-rectangles \cite{pelleg2001mixtures,chen2016interpretable} or polytopes \cite{lawless2021interpretable,gabidolla2022optimal,lawless2023cluster} depending whether the rules delimiting the clusters are axis-parallel or oblique. 

A third family of interpretable clustering techniques is called Descriptive Clustering. It aims at integrating in a same framework distance-based clustering and conceptual clustering. It was first introduced in \cite{dao2018descriptive} and later reused in \cite{davidson2018cluster,sambaturu2020efficient,zhang2021deep}.
This kind of approach considers a framework where data is described by two modalities, one is used to build compact clusters whereas the other (composed of semantic tags) is used to build interpretable descriptions of this cluster. A description is a set of tags and each tag must satisfy some coverage constraints to belong to the description. Clustering is formulated as a bi-objective optimization problem (compact cluster/interpretable cluster). Our work differs strongly from this setting. First, more complex cluster descriptions are considered (patterns, i.e. conjunction of Boolean descriptors instead of tags). Yet, the most important differences are the introduction of the notion of discriminative patterns and its formalization leading to a new setting for interpretable clustering. 

Most of the existing constrained clustering approaches focus on building a partition under the guidance of the constraints. Some approaches first generate a set of partitions and then select the best among them according to their constraint satisfaction \cite{van2017constraint}. To our knowledge, only \cite{mueller2010integer,Ouali2016} focus on the creation of a partition from a given set of clusters while satisfying some user-defined constraints. Our main differences regarding these work are first the formalization of the notion of explainability, the integration of any kind of clustering and conceptual clustering and the consideration of coverage and discrimination constraints. 
ECS can be compared to clustering ensemble methods that generate many base partitions and return a single final partition, often called the consensus partition.
In recent years, multiple approaches aiming at integrating expert knowledge in clustering ensemble methods have been proposed \cite{guilbert2022anchored,yang2012semi,yang2017cluster,wei2018combined,yang2019constraint,yu2014double,yu2017adaptive,yu2018semi}, but none has focused on cluster selection from a set of clusters, nor on interpretability.

Let us notice that although the notions of coverage and discrimination are related to concept learning (see for instance \cite{pami-MichalskiS83} introducing the notions of characteristic and discriminant descriptions), our work differs from concept learning where classes are known and data is already labelled.

\section{Interpretable clustering formulation} \label{sec:pbSta}

Let $\Dts$ be a dataset composed of $\N$ instances. Data is described by a set $\featureSpace$ of features and a set $\tagSpace$ of Boolean descriptors thus composing two views. These two spaces can be the same, overlapping or completely disjoint. The aim is to find a clustering $\Part$ of $\Dts$ where each cluster $C_k$ is computed according to the feature space $\featureSpace$ and is associated to an explanation $\D_k$ composed of one or several patterns
subsets of $\tagSpace$.

Given a pattern $p$ (a set of descriptors) and an instance $\elem$, the predicate $cover$ defines whether $p$ covers $\elem$:
\begin{equation}
    cover(p,o) \defff \forall t\in p, \tagSpace_{o,t}=1 
    \label{eq:coverP}
\end{equation}

A patterns describing a cluster should cover most of its instances ({\em coverage constraint)} and must not cover most of the instances of other clusters ({\em discrimination constraint}).
This is modeled as follows, introducing parameters that will be be set by the expert, allowing the data exploration:

\begin{itemize}
    \item \textbf{Coverage}:
    a pattern $p$ \textit{covers} a cluster $C$ when it covers at least a ratio $\theta\in [0,1]$ of its instances: 
    \begin{equation}
        coverC(p,C,\theta) \defff \#\{o\in C\mid cover(p,o)\}\geq \theta\: \#C
        \label{eq:coverC}
    \end{equation}
    where $\#C$ represents the cardinality of $C$.
    
    \item \textbf{Discrimination}: discriminative explanations must emphasize the differences between instances of a particular cluster and the others. 
    We distinguish 3 levels of discrimination for a pattern $p$ describing a cluster $C$:
    \begin{itemize}
    
    \item \textbf{Clustering-wise}: 
        $p$ should not describe more than a ratio $\eta$ clusters of $\Part$ 
        \begin{equation}
            \#\{ C\in \Part \;|\; coverC(p,C,\theta)\}\; <\eta\: \#\Part
            \label{eq:discrClusteringWiseDescrScale}
        \end{equation}
        This criterion is pretty permissive since depending on $\eta$ this pattern may also cover  other clusters.

        \item \textbf{Dataset-wise}: $p$ should not cover more than a specified ratio $\varrho$ of instances $o\notin C$
        \begin{equation}
            \#\{ o\in \Dts\backslash C \; |\;cover(p,o)\} <\varrho\: \#(\Dts\backslash C)
            \label{eq:discrDatasetWise}
        \end{equation}
  
        \item \textbf{Cluster-wise}: $p$ should not cover more than a certain amount $\phi$ of instances of each cluster $C'\neq C$ 
        \begin{equation}
            \#\{ o\in C' \; |\;cover(p,o)\}\; <\phi\: \#C'
        \end{equation}
        When $\phi = \varrho$, this constraint is stronger than the previous one: if the cluster-wise constraint is satisfied,  the dataset-wise constraint is also satisfied. 
        
    \end{itemize}
    
    Let us notice that the data-set wise discrimination constraint does not depend on the clusters forming the clustering $\Part$. It is a \textit{ClusterVsAll} constraint, considering discrimination in regards of the instances \textit{outside} of the cluster whereas the clustering-wise and the cluster-wise constraints depend on the other clusters of $\Part$. These requirements make the problem combinatorial.
    A explanation $\D_k$ of a cluster $C_k$ is a set of patterns, $\D_k\subseteq \mathcal{P}(\tagSpace)$. It is considered discriminative when \textit{all} of its patterns are discriminative, the notion of discrimination being defined by one or several of the previous definitions.
    
\end{itemize}

\section{Interpretable Cluster Selection} \label{sec:novel}

\subsection{Overview of the approach}

\begin{figure*}[h]
    \centering
    \includegraphics[scale=0.8]{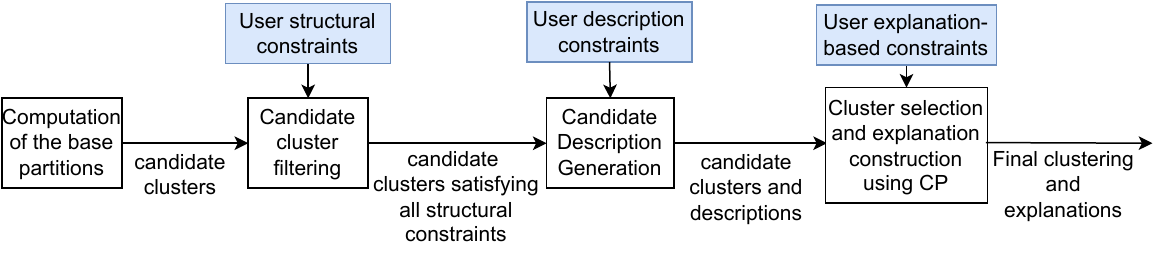}
    \caption{Overall ECS approach}
    \label{fig:overviewHorizontal}
\end{figure*}

The goal is to identify a clustering $\Part$ for dataset $\Dts$ where each cluster $C_k$ is constructed based on the feature space and associated with an explanation $D_k$ composed of covering and discriminative explanations. Our method, based on an ensemble approach, builds a set of clusters and their corresponding covering descriptions. Cluster explanations are sets of discriminant descriptions. 
Because of the discrimination constraints, the choices of clusters and descriptions are highly interdependent. Thus, cluster selection and explanation construction are performed simultaneously.
Our framework allows the integration of expert knowledge expressed as constraints, which may impact cluster structure, such as must-link and cannot-link constraints, or be explanation-based constraints (coverage and discrimination) on explanations, specifying the expert's desired explanation level. Constraints can filter candidate clusters, apply to explanations, or serve as global constraints on the final clustering and explanations. These constraints collectively compose a combinatorial problem.
The main steps of our approach are given in Fig. \ref{fig:overviewHorizontal}.

Given a dataset $\Dts$ of $\N$ instances, different clustering algorithms, e.g. K-Means or hierarchical clustering are applied to generate base partitions, thus leading to a set of $\nbBaseClust$ clusters.
According to some criteria and taking into account user constraints,
a filtering step is performed and for each remaining cluster, a set of patterns is generated. The selected patterns must satisfy the coverage and dataset-wise constraints, if required by the expert.
Finally, a CP model is used to generate a clustering $\Part$ and explanations for each cluster of $\Part$, composed of a subset of its candidate patterns
satisfying the user constraints and the explanation-based constraints.

\subsection{Building candidate clusters}

The generation of candidate clusters is similar to the one in clustering ensemble methods.
Many techniques can be applied with the objective to ensure a diversity in the set of candidate clusters: using the same clustering algorithm with different parameters, applying different algorithms, different data representations, different subsets of the data, projection of the objects in different subspaces.
This step is generic and in Section \ref{sec:exp}, we specify how it is implemented in our experiments. 
It generates many clusters, where some of them may not be worth of interest. A filtering step is thus performed.

\subsection{Cluster filtering}
Different filtering criteria can be applied to remove uninteresting candidate clusters.

\begin{itemize}

    \item Individual cluster selection: constraints on individual clusters are enforced to remove the violating clusters
    \begin{itemize}
        \item Constraints on  clusters, for instance  their sizes, or their diameters.
        \item Expert knowledge expressed by constraints: for instance 
            \textit{Must-Link}$(a,b)$ or
            \textit{Cannot-Link}$(a,b)$.
    \end{itemize} 
     \item Top-k cluster selection: keep the top $\iota\%$ of the clusters according to a specific metric such as the cluster diameter or the Within Cluster Sum of Square (WCSS).
\end{itemize}

\subsection{Generate candidate cluster descriptions} \label{sec:genClustDescr}

For each remaining cluster $C$, the patterns $p$ ($p\subseteq \tagSpace$) candidates for describing $C$ (also called candidate descriptions) are frequent closed patterns appearing in $C$, with the coverage threshold $\theta$. 
The LCM algorithm (Linear time Close itemset Miner) \cite{uno2003lcm} is  applied on each cluster to obtain the list of frequent closed patterns covering at least a ratio $\theta$ of the elements of $C$. The coverage constraint is thus ensured. In order to enforce the dataset-wise constraint, among the generated patterns, only dataset-wise discriminative patterns are kept, i.e. patterns covering less than a ratio $\varrho$ of the instances outside $C$, by Eq. (\ref{eq:discrDatasetWise}).

Once the covering patterns have been computed for each cluster, only the  clusters with at least one pattern are kept.
We obtain  then a set of clusters and explanations that satisfy only the cluster-level and dataset-wise discriminative patterns. Clustering-level and cluster-wise discrimination constraints, which are combinatorial, are enforced in the last step.
\subsection{Generate an explainable clustering}
\label{subsec:partition}

We define the Interpretable Cluster Selection (ICS) problem as: given a dataset $\Dts$ of $\N$ instances and a set of $\nbCand$ candidate clusters
$c$ ($c \subseteq \Dts$) with their set of candidate patterns 
$D_c$, generate an interpretable clustering $\Part$ composed of a subset of the input clusters, and for each selected cluster $C_c$, an explanation $D'_c$ being a subset of their candidate patterns ($D'_c \subseteq D_c$), satisfying the constraints. 
The ICS problem is formulated as a Constrained Optimization Problem (COP) presented in Sec. \ref{sec:model}, which will be solved by a Constraint Programming (CP) solver. Let us recall that a COP is defined by a set of variables with their domains and by a set of constraints. CP solver iterates constraint propagation and search until the best solution satisfying all the constraints is found.

\section{CP model for interpretable cluster selection} \label{sec:model}

This section presents the ICS problem and a Constraint Programming (CP) model. As inputs each candidate cluster $c$ has an associated set of candidate patterns $D_c$ ($D_c \subseteq \mathcal{P}(\tagSpace)$). The aim is selecting some clusters to form a clustering $\Part$, such that each selected cluster $C_c$ is described by some patterns $p\in \D_c$ satisfying the covering and discriminative constraints and the user constraints. In the CP model, the cluster selection and the description selection will be presented by decision variables and the requirements (clustering composition, explanation, etc.) by the constraints of the model.

In order to model this problem, CP \cite{Rossi2006} is chosen rather than Integer Linear Programming (as used for instance in \cite{mueller2010integer}) since many constraints are written as logical implications, which are more naturally expressed in CP. Moreover, non linear constraints on clusters can be added. Compared to SAT, constraints are expressed in a higher language level, making the model much more readable than by decomposing them into CNF formulas.

\subsection{ICS inputs}
        \begin{itemize}
        \item $\N\in\mathbb{N}$: number of data instances.
        \item $\nbPat\in\mathbb{N}$: number of candidate patterns occurring in at least one candidate cluster.
        \item $\nbCand\in\mathbb{N}$: number of candidate clusters.
        \item $\instanceByClust\in \{0,1\}^{\nbCand\times \N}$: a matrix where $\instanceByClust_{ic}=1$ if instance $i$ is in  cluster $c$, and 0 otherwise.
        
        \item $\descrSpace=\{D_1,...,D_\nbCand\}$: each $D_c$ ($D_c\subseteq\mathcal{P}(\tagSpace)$) is a list of candidate patterns for cluster $c$.
        \item $\patternByClust\in \mathbb{N}^{\nbCand\times \nbPat}$: a matrix where $\patternByClust_{cp}$ is the number of instances in cluster $c$ covered by the pattern $p$.
    \end{itemize}
In the following the notation $[1,V]$ denotes the set of the integers $\{1,\ldots,V\}$.
\subsection{Decision variables}
    \begin{itemize}
    \item $\Part\in \{0,1\}^\nbCand$: the final clustering, where $\Part_c=1$ means that cluster $c$ is selected and 0 otherwise.
    \item $\candidateDescr=\{\candidateDescr_1,...,\candidateDescr_\nbCand\}$, each $\candidateDescr_c\in \{0,1\}^{|D_c|}$ and $\candidateDescr_{cp}=1$ means that  pattern $p\in D_c$ is used to explain cluster $c$.
    \item $\nbAppar\in\mathbb{N}^\N$: $\nbAppar_i$ is the number of selected clusters in $\Part$ containing instance $i$. 
    \end{itemize}

\subsection{Constraints}

The model can be defined by the constraints selected by the expert among different constraints below.
They may be on the clustering, on the explanations or user-knowledge constraints.

\paragraph{Clustering constraints}

\begin{itemize}
\item Number of clusters in a given range:
\begin{equation}
    \K_{min} \leq \sum_{c\in [1,\nbCand]} \Part_c \leq \K_{max}
    \label{eq:constrNbClust}
\end{equation}
with $\K_{max}$ (or $\K_{min}$) the maximum (resp. minimum) number of selected clusters.
\item Number of clusters that belongs an instance.
This number can be restricted:
\begin{equation}
    \forall i\in [1,\N],\ nbClustMin\leq \nbAppar_i \leq nbClustMax
    \label{eq:limtNbclust}
\end{equation}
with $\nbAppar_i$ the number of selected clusters that instance $i$ is assigned to, defined by:
\begin{equation}
            \forall i\in [1,\N],\ \nbAppar_i = \sum_{c\in [1,\nbCand]}(\Part_c \times \instanceByClust_{ci})     \label{eq:nbClust(i)}
\end{equation}

Setting $nbClustMin=nbClustMax=1$ will result $\Part$ a hard partition.
If $nbClustMin = 0$, some instances may be unassigned to any cluster in $\Part$.
\item Overlapping control. This constraint sets a bound on the number of instances that are assigned to more than one cluster ( $\mathbf{1}(\cdot)$ is the indicator function).

\begin{equation}
    \sum_{i\in [1,\N]} \mathbf{1}(\nbAppar_i>1) \leq nbDiff1Max
    \label{eq:limtNbDiff1}
\end{equation}

\end{itemize}
\paragraph{Explanation-based constraints} 

Let us recall that coverage and dataset-wise constraints have already been enforced during the filtering step (Subsec. \ref{sec:genClustDescr}). The cluster-wise and clustering-wise explanation conditions are expressed below.
\begin{itemize}
\item Non empty explanation.
All selected clusters must have at least one pattern selected in its explanation.
\begin{equation}
    \forall c\in [1,\nbCand],\; \Part_c=1 \Longleftrightarrow \sum_{p\in [1,|\candidateDescr_{c}|]}\candidateDescr_{cp}\geq 1
    \label{eq:allClustHasDescr}
\end{equation}
\item Clustering-wise constraint. 
A pattern $p$ selected in the final explanation of a selected cluster $c$ should not describe more than a ratio $\eta$ of the selected clusters in $\Part$. 
\begin{multline}
    \forall c\in [1,\nbCand],\ \forall p\in [1,|\candidateDescr_c|], \cr
    \candidateDescr_{cp}=1 \implies \sum_{c'\in [1,\nbCand], \patternByClust_{c'p}> \theta\#c'}\Part_{c'} \leq \eta\sum_{c'\in [1,\nbCand]}\Part_{c'}
    \label{eq:clusteringWiseConstraint}
\end{multline}

\item Cluster-wise constraint.
This constraint requests that a pattern describing a selected cluster $c$ should not cover more than a ratio $\phi$ of another selected cluster $c'$.
Therefore if a pattern $p$ is selected in the final explanation of a selected cluster $c$, then all other cluster for which $p$ covers more than $\phi$ elements cannot be selected.
\begin{multline}
    \forall c\in [1,\nbCand],\ \forall p\in [1,|\candidateDescr_c|], \cr
    \forall c'\in [1,V], c'\neq c\; \mbox{ s.t. }\ \patternByClust_{c'p} > \phi\#c',\cr \candidateDescr_{cp}=1 \implies \Part_{c'}=0
    \label{eq:patClusterWise}
\end{multline}

\item Explanation completeness under the cluster-wise parameter $\phi$. 
If a cluster is selected then all of its patterns that are discriminative towards all other selected cluster are selected in its final explanation.

\begin{multline}
    \forall c\in [1,\nbCand],\ \forall p \in [1,|\candidateDescr_c|], \cr \Part_c=1 \wedge \bigwedge_{c'\in [1,\nbCand], c'\neq c, \patternByClust_{c'p}>\phi\#c'} \Part_{c'}=0\implies  \candidateDescr_{cp} = 1
        \label{eq:enforceSelectionOfAllPossiblePatPhi}
\end{multline}

\end{itemize}

\paragraph{User constraints}

We introduce Must-Select (MS) and Cannot-Select (CS) constraints:
$MS(c)$ requires cluster $c$ to  be selected in the final clustering $\Part$, while $CS(c_i,c_j)$ forbids the simultaneous selection of clusters $c_i$ and $c_j$.
    \begin{itemize}
        \item {\it Must-Select}$(c)$: 
            \begin{equation}
                \Part_c=1
                \label{eq:mustSelect}
            \end{equation}
        \item {\it Cannot-Select}$(c_i,c_j)$:
            \begin{equation}
                \Part_{c_i}+\Part_{c_j}< 2
                \label{eq:cannotSelect}
            \end{equation}
    \end{itemize}
    
 Other constraints could be easily introduced.

\subsection{Objective functions}

Classic objective functions on clustering quality can be used, such as minimizing WCSS or minimizing the diameter.
We propose new objective functions based on explantion quality.

\begin{itemize}
    \item Hard partition quality: maximize the number of instances assigned to one and only one cluster
   
        \begin{equation}
            \arg\max \sum_{i\in [1,\N]}\mathbf{1}(\nbAppar_i=1)
            \label{eq:objNbAppar}
        \end{equation}
    or minimize the number of unassigned instances
        \begin{equation}
            \arg\min \sum_{i\in [1,\N]}\mathbf{1}(\nbAppar_i=0)
            \label{eq:objNbAppar0}
        \end{equation}
    
    \item Maximize the number of selected clusters.
    
        \begin{equation}
            \arg\max \sum_{c\in [1,\nbCand]}\Part_c
            \label{eq:objNumberOfClust}
        \end{equation}
        
    \item Explanation information: maximize the sum of lengths of selected clusters explanations
    
        \begin{equation}
            \arg\max \sum_{c\in [1,\nbCand]}(\Part_c \times |\candidateDescr_c|)
            \label{eq:objMaxSumLengthDescrModel}
        \end{equation}
    
    \item Explanation concision: minimize the sum of length of selected clusters' explanations  
    
        \begin{equation}
            \arg\min \sum_{c\in [1,\nbCand]}(\Part_c \times |\candidateDescr_c|)
            \label{eq:objMinSumLengthDescrModel}
        \end{equation}
        
    \item Minimize the average WCSS of the selected clusters, where $WCSS_c$ represents the sum of the squared distances of instances to the center of the cluster.
    
        \begin{equation}
            \arg\min \sum_{c\in [1,\nbCand]}(\Part_c \times WCSS_c)) \;\slash \sum_{c\in [1,\nbCand]} \Part_c
            \label{eq:objMinAverageWCSS}
        \end{equation}

    \end{itemize}

\section{Experiments} \label{sec:exp}

This section aims at answering the following questions:
\begin{itemize}
    \item Q1: What is the impact of the coverage and discrimination parameters on the results ? (Tables \ref{table:IrisCovDiscrPat}, \ref{table:IrisCovDiscrTag})
    \item Q2: What is the concrete difference between using LCM  patterns  or  single  descriptors? (Tables \ref{table:countryPat}, \ref{table:countryTag})
    \item Q3: What is the interest of allowing unassigned instances? (Section \ref{sec:q3})
    \item Q4: How is ECS wrt. competitors ? (Table \ref{table:Baseline})
\end{itemize}

\subsection{Evaluation metrics}

To evaluate the quality of our results in terms of interpretability, we define three novel evaluation measures for explanations, two for measuring the coverage and one for evaluating the discrimination criterion. All of these values range between 0 and 1, with 1 the most desirable outcome.

\subsubsection{Coverage evaluation metrics}

For a pattern $p$ in the explanation of a cluster $C$, the Pattern Coverage Rate (PCR) is defined by:
\begin{equation}
    PCR(p,C)=\frac{\#\{o\in C : cover(p,o)\}}{\#C}
    \label{eq:PCR}
\end{equation}

We define a second coverage measure, called Explanation Coverage (EC), measuring the ratio of instances that are covered by at least one pattern:
\begin{equation}
    EC(D,C) = \frac{\#\{o\in C \mid \exists p\in D\, cover(p,o)\}}{\#C}
\end{equation}

\subsubsection{Discrimination evaluation metric}

To evaluate the cluster-wise discrimination of a pattern, we define the Inverse Pattern Contrastivity, with $\Part$ being the clustering: 
\begin{equation}
    IPC(p,C_i)=\frac{1}{K-1} \sum_{C'\neq C_i \in \Part} 1-\frac{\#\{o\in C' : cover(p,o)\}}{|C'|}
    \label{eq:ITC}
\end{equation}
It computes the mean on the clusters different from $C_i$ of the percentages of instances not covered by $p$.
The value ranges between 0 and 1 with 1 the best outcome, when $p$ covers no instances outside $C_i$.

\subsection{Experimental Setup}

We performed our experiments on two publicly available UCI datasets (Iris and Flags) and on the image dataset Animal with Attributes 2 (AwA) \cite{AwA2}. For creating the description space, we proceed as follows:

\begin{itemize}
    \item Iris: discretization of numeric attributes into 2 (or 4) intervals according to the median (or quartiles, resp.).
    
    \item Flags: only information related to countries are kept (no information related to their flags), same feature space and description space: landmass (continents), zone based on the intersection between Greenwich and the Equator (NE, SE, SW, NW), area, population, language and religion. Numerical attributes (area, population,\ldots) are discretized into two binary attributes according to the median, categorical attributes with $n$ values are transformed into $n$ Boolean attributes.
    \item AwA2: contains 37322 images of 50 animal classes, each described by a subset of 85 binary tags representing properties such as colors, geographic regions, diets, etc. (descriptor space $\descrSpace$). Each image has been transformed into 200 SWIFT numerical values (feature space $\featureSpace$).  Only the first 10 classes (\textit{antelope, grizzly bear, killer whale, beaver, dalmatian, Persian cat, horse, German shepherd, blue whale} and \textit{Siamese cat}) have been considered, resulting in 7030 instances.  
\end{itemize}

The method has been implemented in Python 3, and the CP model implemented using CPMPY \cite{guns2019increasing}. The code will be made available upon acceptation in an international conference. 
Experiments were performed on a laptop with a 11th Gen Intel® Core™ i7-1165G7 @ 2.80GHz × 8 processor and 31,0 GB memory.

Unless specified otherwise, base partitions are generated with the K-Means algorithm run twice using the Euclidean distance and with $\K$ ranging from 2 to 15.
Patterns are generated with LCM (Linear time Close itemset Miner) algorithm \cite{uno2003lcm} using \cite{pedregosa2011scikit}. 

Let us emphasize that our proposed method is built with expert interaction in mind. 
It is up to the user to define which parameters seems to better fit the task and to modify them according to the results.

Compared to decision tree-based interpretable clustering approaches, our framework allows for user constraints, overlapping and flexibility, whereas these techniques indeed enforce complete separation between nodes, and total coverage of properties separating them. The properties of the explanations that are searched for are thus different.

\subsection{Illustration on an artificial dataset} \label{sec:ArtificialDts}

In this section we study the results obtained on a well-known artificial 2D dataset Halfmoon, containing two crescent moon shapes. This dataset was generated using the $make\_moons$ function in the Python library scikit-learn, with 100 points in each crescent and a noise of 0.05.

A descriptor space was randomly generated by adding color, shape and size properties.
\begin{enumerate}
    \item \textit{Colors}: 80\% of the points in the upper cluster are assigned to the color blue, and the remaining points to the color white. Similarly, for the other cluster 80\% of the points are assigned to red and the remaining ones to white.
    \item \textit{Shapes}: 40\% of all the points are designated as squares, another 40\% as circles and the remaining 20\% are triangles.
    \item \textit{Size}: 50\%  of all the points are designated as small whereas all the others are set to big.
\end{enumerate}

\begin{figure}
    \centering
    \includegraphics[scale=0.5]{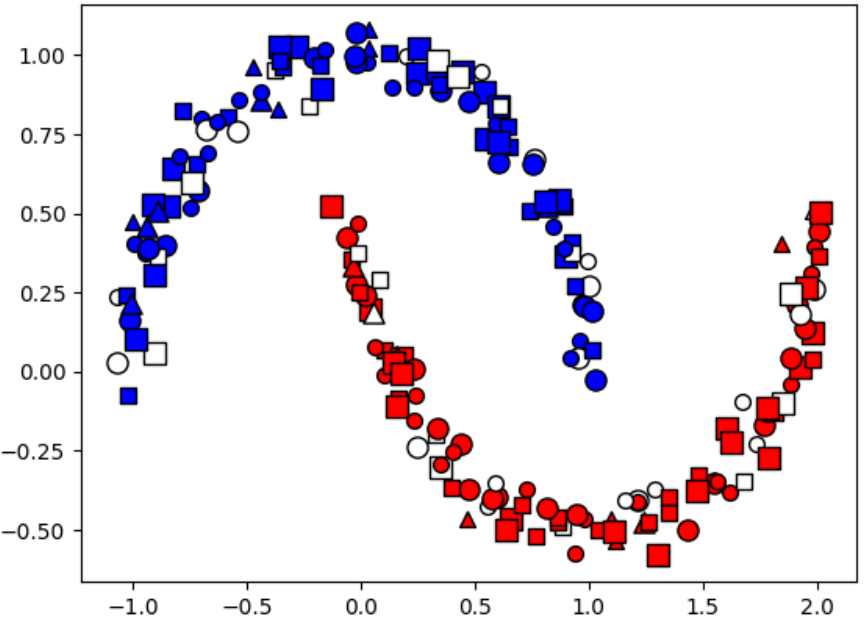}
    \caption{Visual representation of the artificially generated descriptor space on the Halfmoon dataset.}
    \label{fig:moonDescrSpace}
\end{figure}

Fig. \ref{fig:moonDescrSpace} illustrates the descriptor space on the dataset.

In this section, we aim at analyzing the behavior of the ECS method on three aspects: 
(1) the impact of the choice of base clustering algorithms, (2) the ability
to obtain the expected properties, (3) a comparison with the most popular type of decision tree-based approach. 

\subsubsection{Impact of the choice of base clustering algorithms} \label{sec:clustAlgImpact}

\begin{figure}
     \centering
     \begin{subfigure}[b]{0.4\textwidth}
        \centering
        \includegraphics[scale=0.3]{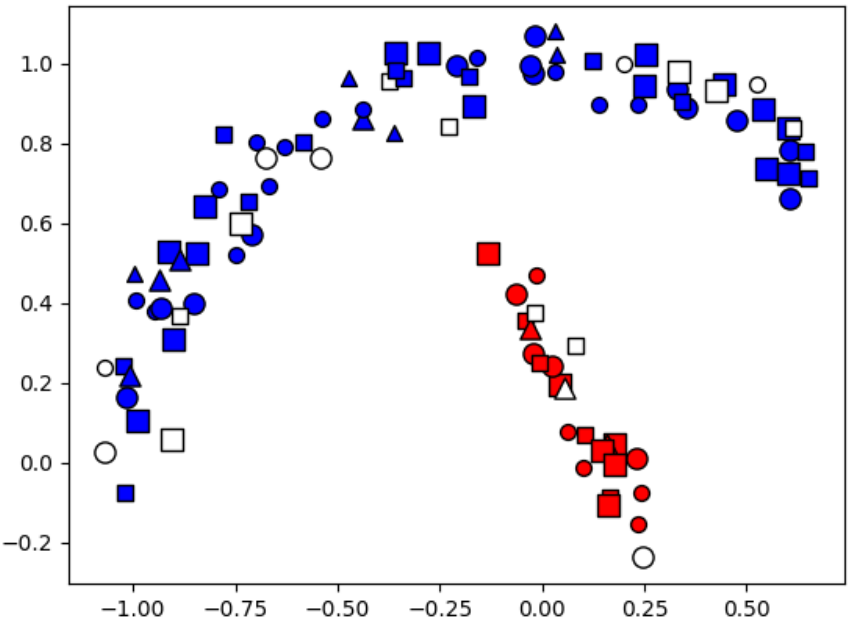}
        \caption{K-Means Cluster 0}
        \label{fig:K-MeansC0}
     \end{subfigure}
     \begin{subfigure}[b]{0.4\textwidth}
        \centering
        \includegraphics[scale=0.3]{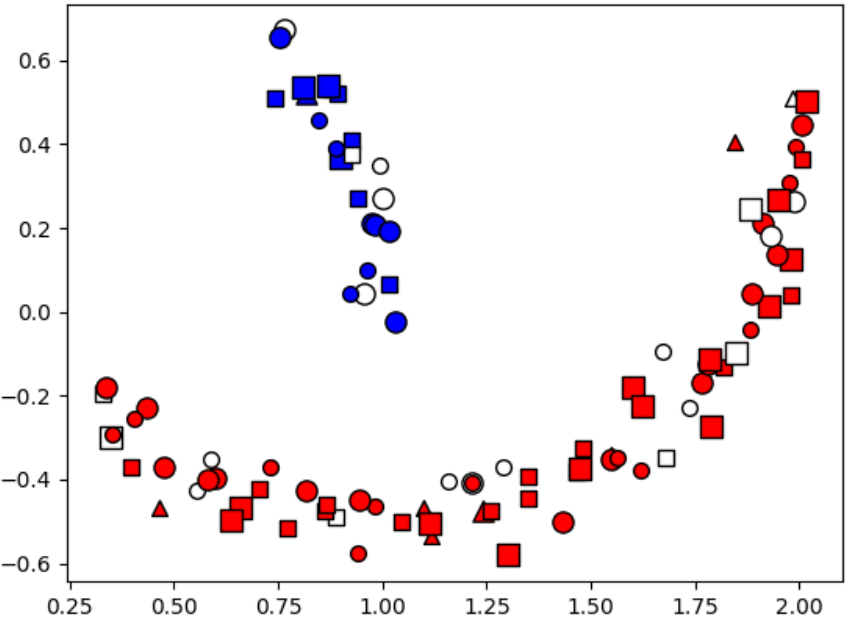}
        \caption{K-Means Cluster 1}
        \label{fig:K-MeansC1}
     \end{subfigure}
    \caption{Example of clusters obtained with K-Means, with ARI=0.26. The explanations with $\theta=60\%$ and $\phi=30\%$ are respectively $\{blue\}$ for cluster 0 and $\{red\}$ for cluster 1.}
    \label{fig:K-MeansMoons}
\end{figure}

We compare three ways of generating base partitions: using K-Means, spectral clustering, or a mix with K-Means and spectral clustering.
When generating base partitions using only K-Means and not allowing overlapping, our system is only able to create partitions with two clusters significantly different from the groundtruth. The Adjusted Rand Index (ARI) \cite{hubert1985comparing}, a measure of the similarity between two partitions, yields a value of 0.26 when comparing the expected partition and the one obtained with K-Means, highlighting the poor quality of these results.
Fig. \ref{fig:K-MeansMoons} illustrates a partition obtained with K-Means, with the first cluster on the left and the second on the right.
With such clusters, it is not possible to find covering and discriminant explanations with a  coverage more than 60\% or a discrimination more than 30\%.

By contrast, when using Spectral Clustering to compute base partitions the ground-truth clusters are obtained (ARI=1.0) (Fig. \ref{fig:moonDescrSpace}). This is true even when clusters are generated both by K-Means and Spectral Clustering.

Different explanations can be obtained depending on the chosen parameters. We now show that ECS can produce explanations containing the expected properties.
(The impact of the coverage and discrimination parameters is explored further in Sec. \ref{sec:CovDiscr}). 

\subsubsection{Validation of expected results}

Because of the method used to generate the Boolean descriptor space, the blue and red colors exclusively appear in their respective clusters, covering 80\% of their points. 
These two properties are thus the most covering and discriminative ones, and are expected to be found in all explanations built by ECS with a coverage parameter set to 80\% or lower. No other property covers more than 60\% of the points in the same cluster.

It is thus not surprising that the explanations that have been found with parameters of coverage and discrimination set to $\theta=70\%$ $\phi=30\%$ for the upper and lower moons are respectively \textit{blue} and \textit{red}, with the same values of coverage and discrimination.

\bigbreak
The random attribution of size and shape attributes results in interesting combinations of properties in terms of coverage and discrimination satisfied by the clusters.
We show that ECS is able to obtain them, depending on the parameters.

For example, we notice that 40$\%$ of both clusters are \textit{big and non-white} (i.e. \textit{big and blue} or \textit{big and red} instances). We can thus expect that conjunctive patterns $\{big,blue\}$ and  $\{big,red\}$ appear in the explanations when setting the coverage parameter equal to or below 40. 
The attribute \textit{big} in itself covers resp. 51 objects in the upper cluster and 49 objects in the other. While putting a coverage parameter below 50\% could allow it to be in the explanation of both clusters, the discrimination constraint
(if restrictive enough) should discard it.

Looking at the results of ECS with coverage $\theta=40\%$ and discrimination $\phi=30\%$, we notice that all the expected properties are fulfilled in the final cluster explanations: $\D_0=\{\{blue\}, \{blue, big\}\}$ and $\D_1=\{\{red\}, \{red, big\}\}$.

\subsubsection{Comparison to decision trees-based method}

We now discuss differences and advantages of ECS compared to one of the most popular decision-tree-based explainable clustering algorithm IMM \cite{dasgupta2020explainable}, which is the foundation of most of the recent articles \cite{frost2020exkmc,gamlath2021nearly,esfandiari2022almost,charikar2022near,laber2021price,makarychev2021explainable,makarychev2021near,laber2023shallow,bandyapadhyay2023find}.

IMM, along with their direct extensions, constructs a decision tree with the objective of approximating a K-Means partition. This is the first major difference compared to ECS since, as shown for example in Sec. \ref{sec:clustAlgImpact}, the choice of the algorithm has a huge impact on clustering quality. In the case of the Halfmoon dataset, the fact that IMM is based on K-Means implies that it is unable to find a partition close to the ground-truth. IMM indeed returns partitions obtained with K-Means and the results and cluster composition are thus similar to the one we presented in Sec. \ref{sec:clustAlgImpact} and Table \ref{fig:K-MeansMoons} with ARI=0.26.

Another key difference is that the features used for splitting the tree are the same as those used to compute the original partition. One of our major aspect is indeed the combination of a feature space used to compute the clusters, and a binary descriptor space used for explanations.
This means that, in the context of the Halfmoon dataset, IMM is unable to find explanations based on attributes such as the color or the shapes.

\subsection{Experimental Results} \label{sec:ExpRes}

\subsubsection{Q1: Impact of the coverage $\theta$ and discrimination $\phi$} \label{sec:CovDiscr}

\begin{table}
\centering
\begin{tabular}{c|c|c|c|c|c}
Coverage & Discr. & PCR  & EC   & IPC & Time (sec) \\
\hline
70       & 10     &   -   &   -   &   -   & 0.24 \\
70       & 30     & 0.82 & 0.91 & 0.93 & 4.47\\
70       & 50     & 0.83 & 0.91 & 0.92 & 4.48\\
50       & 10     &   -   &   -   &   -   & 0.12 \\
50       & 30     & 0.69 & 0.97 & 0.95 & 10.87\\
30       & 10     & 0.54 & 0.98 & 0.99 & 22.27
\end{tabular}
\caption{LCM patterns: comparison of the impact of Coverage and Discrimination parameters on Iris datasets. }
\label{table:IrisCovDiscrPat}
\end{table}

\begin{table}
\centering
\begin{tabular}{c|c|c|c|c|c}
Coverage & Discr. & PCR  & EC   & IPC & Time (sec) \\
\hline
70       & 10     &  -   &  -   &  - & 0.08 \\
70       & 30     & 0.83 & 0.91 & 0.91 & 5.07\\
70       & 50     & 0.85 & 0.91 & 0.88 & 4.64\\
50       & 10     &   -   &   -   &   -   & 0.06\\
50       & 30     & 0.73 & 0.97 & 0.91 & 7.07\\
30       & 10     & 0.7  & 0.79 & 0.97 & 9.84 
\end{tabular}
\caption{Single descriptor: comparison of the impact of Coverage and Discrimination parameters on Iris.}
\label{table:IrisCovDiscrTag}
\end{table}

Tables \ref{table:IrisCovDiscrPat} and \ref{table:IrisCovDiscrTag} present results obtained on Iris with respectively explanations composed of LCM patterns or single descriptors.
The base partitions were obtained using K-Means with a number of clusters between 2 and 8 in 0.22 seconds. No overlapping or unassigned instances were allowed and the number of clusters of the final clustering was set to 3. The objective criterion was to minimize WCSS.
Each row represents a different combination of the Coverage and Discriminative cluster-wise parameters $\theta$ and $\phi$. 

The values in the Coverage and Discr. columns are percentages; PCR, EC and IPC columns display the mean of those measures for every cluster.
The last column contains the execution time in seconds of the CP model. 
The character - means that no possible solution were found.

It can be observed that: (1) As expected, it is difficult to find results with both strong coverage and discrimination capacities.
(2) With a low coverage rate, LCM can find more patterns therefore enabling the creation of explanations (set of patterns) covering more instances.
The low PCR values can be explained by the multiplication of patterns with low coverage in the explanations.
(3) Runtime increases when the coverage parameter decreases and depending on the pattern type (LCM or Single descriptor). Decreasing the coverage rate indeed raises the number of candidate patterns, with a negative effect on the complexity of the combinatorial model.
The same remark holds when using LCM, which leads to an exponential number of the candidate patterns.

\begin{table}
\centering
\begin{tabular}{c|c|c|c|c|c}
Coverage & Discr. & PCR  & EC   & IPC  & Clusters \\
\hline
30       & 10     & 0.81 & 0.87 & 1.0  & 5     \\
50       & 10     & 0.77 & 0.83 & 0.99 & 4     \\
50       & 30     & 0.74 & 0.85 & 0.97 & 4     \\
70       & 10     & 0.8  & 0.88 & 0.98 & 2     \\
70       & 30     & 0.84 & 1.0  & 0.89 & 2     \\
70       & 50     & 0.94 & 0.95 & 0.93 & 5     \\
90       & 10     &  -   &   -   &  -    &    -   \\
90       & 30     & 0.92 & 0.92 & 0.8  & 2     \\
90       & 50     & 0.94 & 0.96 & 0.71 & 2    
\end{tabular}
\caption{Results on AwA2 with single descriptors, no overlapping and 15 unassigned instances.}
\label{table:AwAtag}
\end{table}

\begin{table}
\centering
\begin{tabular}{c|c|c|c|c|c}
Coverage & Discr. & PCR  & EC   & IPC  & Clusters \\
\hline
30       & 10   & 0.87 & 0.97 & 1.0  & 7     \\
50       & 10   & 0.91 & 0.97 & 1.0  & 7     \\
50       & 30   & 0.91 & 0.98 & 1.0  & 7     \\
70       & 10   & 0.96 & 0.97 & 1.0  & 7     \\
70       & 30   & 0.96 & 0.98 & 1.0  & 7     \\
70       & 50   & 0.96 & 0.98 & 0.98 & 7     \\
90       & 10   & 0.98 & 0.98 & 1.0  & 7     \\
90       & 30   & 0.97 & 0.98 & 0.99 & 7     \\
90       & 50   & 0.97 & 0.98 & 0.98 & 7    
\end{tabular}
\caption{AwA with LCM patterns.}
\label{table:AwApat}
\end{table}

Table \ref{table:AwAtag} presents results obtained with AwA2 when maximizing the number of clusters. This objective is chosen to demonstrate the impact of parameter choices on the number of clusters (last column) that can be found in $\Part$. 
It can be noticed that different coverage and discrimination parameters can lead to different number of clusters in the final output.

\begin{table}
\centering
\begin{tabular}{ c|p{9.5cm}|c|c|c|c }
C & Explanation & Size & PCR & EC & IPC\\
\hline
0  & \{sw\_m1\} & 62 & 0.73 & 0.73 & 0.85 \\
\hline
1  & \{pl4, sl\_m2, pl\_m2, pw\_m2\}, \{pl4, pw4, sl\_m2, pl\_m2, pw\_m2\}, \{sl4, sl\_m2, pl\_m2, pw\_m2\}, \{sl4, pl4, sl\_m2, pl\_m2, pw\_m2\}, \{sl4, pl4, pw4, sl\_m2, pl\_m2, pw\_m2\}, \{sl\_m2, sw\_m2, pl\_m2, pw\_m2\} & 38 & 0.84 & 1.0 & 0.96 \\
\hline
2  & \{pl\_m1, pw\_m1\}, \{sl\_m1, pl\_m1, pw\_m1\}, \{sw\_m2, pl\_m1, pw\_m1\}, \{sl\_m1, sw\_m2, pl\_m1, pw\_m1\}, \{pl1, pl\_m1, pw\_m1\}, \{pl1, sl\_m1, pl\_m1, pw\_m1\} & 50 & 0.89 & 1.0 & 0.97 
\end{tabular}
\caption{Results on Iris with LCM patterns.}
\label{table:IrisPat}
\end{table}

\begin{table}
\centering
\begin{tabular}{ c|p{3.8cm}|c|c|c|c }
C & Explanation & Size & PCR & EC & IPC\\
\hline
0  & \{sw\_m1\} & 62 & 0.73 & 0.73 & 0.85 \\
\hline
1  & \{sl4\}, \{pl4\}, \{pw4\} & 38 & 0.9 & 1.0 & 0.93 \\
\hline
2  & \{pl1\}, \{pw\_m1\} & 50 & 0.87 & 1.0 & 0.94 
\end{tabular}
\caption{Results on Iris with single descriptors as patterns.}
\label{table:IrisTag}
\end{table}

As for Iris, Tables \ref{table:IrisPat} and \ref{table:IrisTag} contain results respectively with explanations composed of LCM patterns or single descriptors.
By comparing them, the following properties can be observed:
\begin{itemize}
    \item The use of patterns allows to find a solution with $\theta=90\%$ and $\phi=10\%$, whereas it is not possible with single descriptors.
    \item The use of single descriptors does not allow to find many clusters in the final clustering compared to LCM patterns, and it is worse when raising the coverage requirement.
    This shows that single descriptors are not always sufficient to get covering and discriminative explanations, and demonstrates the interest of using patterns.
\end{itemize}

\subsubsection{Q2: LCM patterns or single descriptors.} \label{sec:q2}

Examples of explanations of final clusterings are provided, illustrating the differences between patterns or single descriptors.
Results are presented in Tables \ref{table:countryPat} and \ref{table:countryTag}. The runtime of the CP model to produce these results are respectvly 22.12 and 30.57 seconds.
The base partitions were generated with K-Means with K between 2 and 15. The results are both obtained with the coverage parameter set to 90\% and cluster-wise discrimination to 50\%, allowing for up to 15 unassigned instances and asking for a final clustering with a number of clusters between 2 and 6.
Both results display 6 clusters with similar instance distributions and explanations. 

Let us recall that ECS finds at the same time a clustering and its explanations. We can observe that allowing different sizes of explanations (LCM patterns vs. single tags) the cluster composition may change. From Tables \ref{table:countryPat} and \ref{table:countryTag}, we can see that the two versions agree on the clusters 3, 4 and 5. 
As the utilization of LCM patterns permits the inclusion of covering patterns with multiple descriptors, the composition of clusters 0, 1, and 2 varies, facilitating the reach of higher covering and discrimination scores. Indeed, the integration of LCM patterns results in more discriminant explanations (IPC) and greater coverage (EC) compared to using single descriptors alone.

Looking at Table \ref{table:countryPat}, we notice that cluster 0 has two patterns in its explanation: \textit{Europe} and \textit{Europe \& NE}. While the second has by definition a smaller (or equal) coverage than simply \textit{Europe}, it is more discriminative since some European countries such as Ireland and Iceland are attributed to Cluster 2 which itself mostly contains north-western island nations. This could not be found by considering a single Boolean descriptor.

Cluster 4, containing most Africa countries, is the same in both tables, but its explanation is slightly more discriminative in Table \ref{table:countryTag}. It is because of Sao-Tome (an African country) that is miss-classified in the European cluster in Table \ref{table:countryPat} and unassigned in the clustering in Table \ref{table:countryTag}.

Cluster 1 in Table \ref{table:countryPat} cannot be described by the descriptor \textit{Spanish} because it contains French-Guiana, Haiti and Brazil, making its coverage less then 90\%, whereas in Table 5 it contains only Brazil as a non-spanish speaking countries.

In both clusterings, the clusters numbered 5 contains the same points. The pattern $\{Asia, NE\}$ in Table \ref{table:countryPat} has been selected instead of simply $\{Asia\}$ due to the fact that LCM keeps only closed patterns.

\begin{table}
\centering
\begin{tabular}{ c|p{3.8cm}|c|c|c|c }
C & Explanation & Size & PCR & EC & IPC\\
\hline
0  &\{Europe\}, \{Europe, NE\} & 29 & 0.98 & 1.0 & 0.98 \\
\hline
1  & \{Catholic\} & 24 & 0.96 & 1.0 & 0.89 \\
\hline
2  & \{small pop, NW\}, \{small area, small pop, NW\} & 27 & 0.98 & 1.0 & 0.96\\
\hline
3  & \{Oceania\} & 19 & 0.95 & 0.95 & 0.99 \\
\hline
4  & \{Africa\} &  41 & 0.95 & 0.95 & 0.99\\
\hline
5  & \{Asia, NE\} & 39 & 1.0 & 1.0 & 1.0
\end{tabular}
\caption{Results on Flags with LCM patterns.}
\label{table:countryPat}
\end{table}

\begin{table}
\centering
\begin{tabular}{ c|p{3.8cm}|c|c|c|c }
C & Explanation & Size & PCR & EC & IPC\\
\hline
0  & \{Europe\} & 31 & 0.97 & 0.97 & 0.96 \\
\hline
1  & \{Spanish\}, \{Catholic\} & 15 & 0.93 & 1.0 & 0.92 \\
\hline
2  & \{NW\} & 34 & 1.0 & 1.0 & 0.86 \\
\hline
3  & \{Oceania\} & 19 & 0.95 & 0.95 & 0.99 \\
\hline
4  & \{Africa\} &  41 & 0.95 & 0.95 & 0.97 \\
\hline
5  & \{Asia\} & 39 & 1.0 & 1.0 & 1.0
\end{tabular}
\caption{Results on Flags with patterns of size 1.}
\label{table:countryTag}
\end{table}

\subsubsection{Q3: Interest of allowing unassigned instances.} \label{sec:q3}

Table \ref{table:IrisCovDiscrTag} shows that there is no partition satisfying all the required constraints with $\theta=50\%$ and $\phi=10\%$. However, when allowing 10 unassigned instances, a clustering was obtained with PCR=0.68, EC=0.85 and IPC=0.99.
This shows that allowing unassigned instances can lead to find an interpretable clustering, when a complete partition of the data satisfying the explanation-based constraints does not exist.

\subsubsection{Q4: How is ECS compared to competitors?} \label{sec:q4}

\begin{table}
\centering
\begin{tabular}{c|c||c||c||c}
\multirow{2}{*}{}                & Dataset & Iris & flags & AwA2 \\ \cline{2-5} 
                                 & K       & 3    & 6     & 2   \\ \hline
\multirow{3}{*}{Dao \textit{et al.} 2018} & PCR     & \textbf{1.0}  & \textbf{1.0}   & \textbf{1.0}\\
                                 & EC      & \textbf{1.0}  & \textbf{1.0}   & \textbf{1.0} \\
                                 & IPC     & 0.77 & 0.62  & 0.38  \\ \hline
\multirow{3}{*}{K-Means}         & PCR     & 0.86 & 0.96  & 0.90  \\
                                 & EC      & 0.99 & 0.99  & \textbf{1.0}   \\
                                 & IPC     & 0.73 & 0.70  & 0.39\\ \hline                             
\multirow{3}{*}{K-Means (LCM)}   & PCR     & 0.83 & 0.86 & 0.87  \\
                                 & EC      & 0.99 & \textbf{1.0}  & \textbf{1.0 }  \\
                                 & IPC     & 0.86 & 0.86 & 0.52 \\ \hline                               
\multirow{3}{*}{ECS}  & PCR     & 0.83 & 0.97 & 0.84  \\
                                 & EC      & 0.91 & 0.95 & \textbf{1.0} \\
                                 & IPC     & 0.88 & \textbf{0.95} & \textbf{0.89} \\ \hline                              
\multirow{3}{*}{\begin{tabular}[c]{@{}l@{}}ECS (LCM)\end{tabular}} 
                                & PCR & 0.82 & 0.97 & 0.84 \\
                                & EC  & 0.91 & 0.98 & \textbf{1.0} \\
                                & IPC & \textbf{0.93} & \textbf{0.96} & \textbf{0.89}
\end{tabular}
\caption{Results with naive K-Means baseline and Dao \textit{et al.}, 2018 compared to ECS.
}
\label{table:Baseline}
\end{table}

ECS is compared to:
\begin{itemize}
    \item A bi-objective clustering approach \cite{dao2018descriptive} that aims at finding a Pareto front of partitions maximizing the diameter and the MMCTA criterion (number of tags shared by all instances in a cluster). Since ECS focuses on the quality of the explanations, we consider the partition \v{that the bi-objective clustering approach obtains - Est-ce bien cela ?} with the highest MMCTA. This method considers only individual descriptors (single tag) and is not able to consider patterns.
    \item A baseline that builds a partition using K-Means and  computes for each cluster frequent single descriptors or frequent patterns using LCM, thus integrating no discriminative constraints. 
    The results for each configuration are the average obtained on 5 different K-Means runs.
\end{itemize}

To be fair towards \cite{dao2018descriptive}, we compare against ECS results obtained with single descriptors.
The coverage parameter for ECS and K-Means baseline was set to $70\%$, and the discrimination rate in the ICS model was fixed to $30\%$.

Results are given in Table \ref{table:Baseline}. 
It can first be noticed that \cite{dao2018descriptive} descriptions ensure maximum PCR and EC but are less discriminative than our model.
This is not an unexpected property. Indeed, the high coverage values are due to the requirement put by MMCTA.
The K-Means baseline resulted in clusterings, which have similar or better IPC than \cite{dao2018descriptive} but with inferior pattern coverage.
Overall, in nearly every context ECS generates more discriminative explanations with only small decreases in the coverage measures, which was our goal.

\subsection{Examples of explanations} \label{sec:SupplemExample}

Results with Flags were presented at the previous section. We now introduce examples of explanations generated by our framework with the Iris and AwA2 datasets.

\subsubsection{Explanation for Iris dataset}

Results obtained with Iris are presented in Tables \ref{table:IrisPat} and \ref{table:IrisTag}. Coverage was set to 70 $\%$ and discrimination to 30 $\%$.

The Iris dataset is composed of 150 instances and 4 attributes: sepal length (sp), sepal width (sw), petal length (pl) and petal width (pw).
The descriptor space was created by performing both discretization of numeric attributes into 2 intervals according to the median and a discretization into 4 intervals according to the quartiles, thus leading to 6 Boolean descriptors for each numeric attribute. In the explanations presented in this section, features obtained by discretization according to the median are abbreviated by $m1$ (resp. $m2$) depending on whether it is inferior (resp. superior) to the median, and features obtained by discretization according to quartiles are simply denoted by $1, 2, 3$ or $4$.

For instance in Table \ref{table:IrisTag} cluster $C_0$ is composed of 62 instances and 73$\%$ of these instances satisfy the property to have a sepal-width inferior to the median. On the other hand, cluster $C_1$ is described by 3 patterns involving 3 attributes:
sepal length, petal length and petal width in the 4th quartile. All the instances are covered ($EC=1$), the average coverage rate of the patterns is 90$\%$ and the average discriminativeness is high ($mean(IPC)=0.93$).

Cluster 1 in Table \ref{table:IrisPat} is described by 6 patterns themselves composed of 4 or 5 tags.

We can notice that the explanations of clusters 1 and 2 in Table \ref{table:IrisPat} can be considered as containing too many patterns to be interpretable. Future work will explore the possibility of limiting the number of elements (patterns) in cluster explanations. Let us emphasize that the length of the patterns can be easily set with a threshold and patterns longer than this threshold can be removed when building the explanations of the clusters.

\subsubsection{Explanation for AwA2 dataset}

Tables \ref{table:AwA2ResSupl} and \ref{table:AwA7ResSupl} both present an example of results obtained on the AwA2 dataset with $\theta=90\%$ and $\phi=30\%$ with explanations built respectively with single descriptors, or patterns. Both clustering detailed in those tables also appear in Tables \ref{table:AwAtag} and \ref{table:AwApat}.

Single descriptors only allowed the separation of animals in two clusters depending on whether they live solitary, while using patterns allows more clusters to be selected.

\begin{table}
\centering
\begin{tabular}{ c|p{3.8cm}|c|c|c|c }
C & Explanation & Size & PCR & EC & IPC\\
\hline
0  & \{group\} & 3324 & 0.94 & 0.94 & 0.8 \\
\hline
1  & \{solitary\} & 3706 & 0.91 & 0.91 & 0.8
\end{tabular}
\caption{A result obtained for AwA2 with Single descriptors.}
\label{table:AwA2ResSupl}
\end{table}

\begin{table}
\centering
\scalebox{0.9}{
\begin{tabular}{ c|p{13cm}|c|c|c|c}
C & Explanation & Size & PCR & EC & IPC \\
\hline
0  & \{furry, toughskin, big, lean, hooves, longleg, tail, chewteeth, horns, walks, fast, strong, muscle, quadrapedal, active, agility, vegetation, forager, grazer, newworld, oldworld, plains, fields, mountains, ground, timid, group\} & 1069 & 0.96 & 0.96 & 1.0 \\
\hline
1  & \{black, white, brown, gray, patches, furry, toughskin, big, lean, hooves, longleg, longneck, tail, chewteeth, buckteeth, smelly, walks, fast, strong, muscle, quadrapedal, active, agility, vegetation, grazer, newworld, oldworld, plains, fields, ground, timid, smart, group, domestic\} & 1600 & 0.98 & 0.98 & 0.99\\
\hline
2  & \{brown, furry, toughskin, bulbous, paws, chewteeth, claws, fast, strong, muscle, quadrapedal, active, nocturnal, hibernate, fish, newworld, ground, smart, solitary\} & 1049 & 0.97 & 0.97 & 0.99\\ 
\hline
3  & \{black, white, patches, spots, furry, hairless, big, lean, paws, longleg, tail, chewteeth, meatteeth, walks, fast, strong, muscle, quadrapedal, active, agility, meat, newworld, oldworld, ground, timid, smart, group, solitary, domestic\} & 539 & 1.0 & 1.0 & 1.0\\
\hline
4 & \{black, brown, gray, patches, furry, big, lean, pads, paws, longleg, tail, chewteeth, meatteeth, claws, smelly, walks, fast, strong, muscle, quadrapedal, active, agility, meat, hunter, stalker, newworld, oldworld, plains, ground, fierce, smart, solitary, domestic\} & 1070 & 0.94 & 0.94 & 1.0\\
\hline
5 & \{white, gray, furry, small, pads, paws, tail, chewteeth, meatteeth, claws, walks, fast, weak, quadrapedal, inactive, agility, fish, meat, newworld, oldworld, ground, timid, smart, solitary, domestic\} & 1216 & 0.99 & 0.99 & 0.99\\
\hline
6 & \{toughskin, bulbous, tail, strainteeth, swims, fast, strong, fish, water, smart, group\}, \{spots, hairless, toughskin, big, bulbous, flippers, tail, strainteeth, swims, fast, strong, fish, plankton, arctic, ocean, water, smart, group\} & 486 & 0.97 & 0.99 & 0.99
\end{tabular}
}
\caption{A result obtained for AwA2 with patterns.}
\label{table:AwA7ResSupl}
\end{table}

\newpage
\section{Conclusion} \label{sec:conclu}

We present a novel constrained clustering ensemble approach called ECS that computes a set of clusters with their explanations, where an explanation is a set of patterns covering most of the instances of the cluster and discriminant w.r.t. the other clusters. These notions of coverage and discrimination are defined by means of constraints and parameters to set on the expected explanations of the clusters. Two representation spaces are needed, one for clustering and the other one for explanations. The latter can be computed from the first one thanks to discretization or other preprocessing methods.   
We have applied our method to different datasets and configurations, and show that our system is able to build clustering with interesting explanations. We see our method as a tool for interactive clustering, where the expert can set the parameters depending on his/her expectations and on the results of previous clusterings.
Further directions will be studied: optimization of the CP model so that it can handle a larger number of clusters and patterns, introduction of new constraints on the explanations as the concision or the diversity of the patterns.

\section*{Acknowledgments}

This work was funded by the ANR project InvolvD (Interactive constraint elicitation for unsupervised and semi-supervised data mining) (ANR-20-CE23-0023).

\newpage
\bibliographystyle{named}
\bibliography{references.bib}

\end{document}